\documentclass[11pt,a4paper]{article}
\usepackage[hyperref]{acl2020}
\usepackage{times}
\usepackage{latexsym}

\usepackage{microtype}

\aclfinalcopy 

\usepackage{times}
\usepackage{graphicx}
\usepackage{latexsym}
\usepackage{amsfonts}
\usepackage{amsmath}
\newcommand{\mathbbm}[1]{\text{\usefont{U}{bbm}{m}{n}#1}}
\usepackage{arydshln}
\usepackage{subcaption}
\usepackage[]{xcolor}
\usepackage{ulem}
\title{{VAE based Text Style Transfer with Pivot Words Enhancement Learning}}

\newcommand\Mark[1]{\textsuperscript#1}
\newcommand\extrafootertext[1]{%
    \bgroup
    \renewcommand\thefootnote{\fnsymbol{footnote}}%
    \renewcommand\thempfootnote{\fnsymbol{mpfootnote}}%
    \footnotetext[0]{#1}%
    \egroup
}

\author{
    Haoran Xu \Mark{1}\Mark{,}\Mark{2},
    Sixing Lu \Mark{2},
    Zhongkai Sun \Mark{2},
    Chengyuan Ma \Mark{2},
    Chenlei Guo \Mark{2}\\
    \Mark{1}Johns Hopkins University, \Mark{2}Amazon Alexa AI \\[1em]
    \texttt{hxu64@jhu.edu}\\
    \texttt{\{cynthilu, zhongkas, mchengyu, guochenl\}@amazon.com}
    
}

\date{}

\begin{document}
\maketitle
\extrafootertext{Work done during an internship at Amazon Alexa AI.}
\begin{abstract}
Text Style Transfer (TST) aims to alter the underlying style of the source text to another specific style while keeping the same content. Due to the scarcity of high-quality parallel training data, unsupervised learning has become a trending direction for TST tasks. In this paper, we propose a novel VAE based Text Style Transfer with pivOt Words Enhancement leaRning (VT-STOWER) method which utilizes Variational AutoEncoder (VAE) and external style embeddings to learn semantics and style distribution jointly. Additionally, we introduce pivot words learning, which is applied to learn decisive words for a specific style and thereby further improve the overall performance of the style transfer. The proposed VT-STOWER can be scaled to different TST scenarios given very limited and non-parallel training data with a novel and flexible style strength control mechanism. Experiments demonstrate that the VT-STOWER outperforms the state-of-the-art on sentiment, formality, and code-switching TST tasks \footnote{The code is available at \url{https://github.com/fe1ixxu/VT-STOWER}.}.
\end{abstract}

\section{Introduction}
\begin{figure}[ht]
    \centering
    \includegraphics[width=0.48\textwidth]{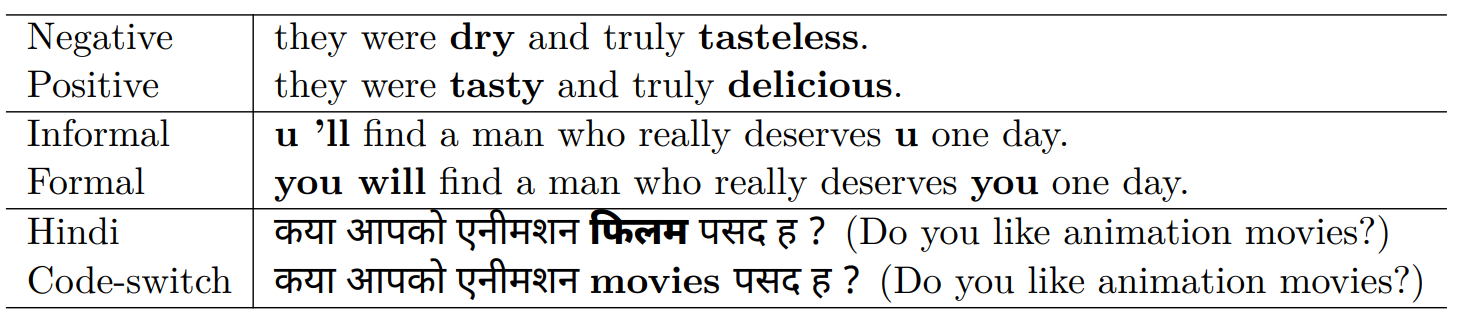}
    \caption{Examples of different TST. Including sentiment style transfer (negative $\leftrightarrow$ positive), formality style transfer (informal $\leftrightarrow$ formal), code-switch style transfer (single language $\leftrightarrow$ code-switch sentence).}
    \label{fig:intro}
\end{figure}

Text style transfer (TST) is an important task in the natural language generation area, aiming to control the certain manner of the semantics style expressed in the generated text. Such styles include but not limit to emotion, humor, politeness, formality, and code-switching. For instance, sentiment transfer is widely seen in sentiment analysis for reviewing comments (e.g., yelp, twitter), and targets on converting the original negative/positive comment into a new comment with same topic but opposite sentiment \citep{hu2017toward,shenstyle}; formality transfer is commonly used in documenting, aims at transferring the informal oral expression into a formal written expression \citep{jin2020deep}.
In this paper, we also consider code-switching as a style transfer task, which has not been explored by previous works. Code-switching is a complicated linguistic phenomenon where a speaker alternates between two or more languages in one utterance, either inter-sentential or intra-sentential. The code-switching transfer is a more challenging task considering cross-lingual alignment and limited available training data in nature. Examples of these three style transfer tasks are shown in the Figure \ref{fig:intro}.

Because of the scarcity of high-quality parallel training data, unsupervised learning has become the mainstream for TST tasks.  Existing works on unsupervised TST learning can be roughly categorized into \textit{Disentanglement} \citep{shenstyle,hu2017toward,fu2018style,john-etal-2019-disentangled} and \textit{Style Attribute Rewriting} \citep{lample2018multipleattribute,dai-etal-2019-style, ijcai2020-526}. Disentanglement approaches strip style features from the content and incorporate the content features with the target style representation. However, researchers become less focus on disentanglement methods after \citet{locatello2019challenging} theoretically proved disentanglement approaches are impossible to represent style fully with unsupervised learning. 
The style attribute rewriting enforces the model to focus on style-independent words by cycle reconstruction and rewriting the style attributes with style embeddings. \citet{dai-etal-2019-style} firstly proposed style transfer model based on the transformer architecture along with target style information. \citet{lample2018multipleattribute} reported that a good decoder can generate the text with the desired style by rewriting the original style. However, the style strength of the generated sentences cannot be easily adjusted in above mentioned works. 

Variational autoencoder (VAE) is firstly proposed by \citep{kingma2013auto} for generation by formatting the latent distribution instead of feeding a single latent feature to the decoder. Many TST models have been benefited from the architecture of VAE. \citet{bowman-etal-2016-generating, john-etal-2019-disentangled} showed that the latent space learned by VAE is considerably smoother and more continuous than the one learned by Deterministic Autoencoder (DAE). \citet{hu2017toward} proposed a new neural generative model that combines VAE and holistic attribute discriminators for effective imposition of the style semantic structures.
In this paper, we also leverage the VAE and propose a novel method called VAE based Text Style Transfer with pivOt Words Enhancement leaRn-ing (VT-STOWER) for TST tasks. VT-STOWER utilizes both VAE and style embeddings to jointly learn the distribution of content and style features. More importantly, we boost the performance of TST tasks much more by inventing pivot words enhancement learning. Compared with other style-transfer methods, our proposed VT-STOWER has a bunch of advantages. In general, the advantages and contributions of the VT-STOWER can be summarised as follows:
\begin{itemize}
\item VT-STOWER integrates the advantages of both VAE and style embeddings. The former catches continuous style expression distribution in language itself while the latter differentiates embedding between original style and target style.
\item VT-STOWER has the flexibility to adjust the target style strength by granting different weights to the auxiliary target style embedding; This allows VT-STOWER to better migrate to different style transfer scenarios, which is rarely studied in previous style-transfer work.
\item With the \textit{pivot words masking enhancement} mechanism, VT-STOWER is able to focus more on the pivot words (certain words that can determine the style of the sentence) and be aware of which words have higher probability to be transferred in the TST. This enhancement significantly improves the transfer accuracy while maintaining original topic.
\item VT-STOWER can be easily scaled to different types of TST tasks. To the best of our knowledge, we are the first to consider code-switching in perspective of style transfer and demonstrate that VT-STOWER can be successfully applied to the Hindi-Hinglish code-switching transfer. Therefore, we provide more potential solutions for the the code-switching problems beyond translation by which translating from single language to code-switching expression is very hard given limited training data.
\item We evaluate VT-STOWER on the benchmark dataset of sentiment, formality transfer tasks, and the code-switching style transfer. Experimental results on all tasks demonstrate better overall performance against state-of-the-art methods, which highlights effectiveness and wide application of VT-STOWER.
\end{itemize}

\section{Proposed Method}

The training of VT-STOWER consists of two stages. The training stage I is a VAE reconstruction task in which the input text $x$ will be reconstructed together with external style embeddings. The latent space of content distribution is learned by VAE, and the original and target style mapping will be learned and saved in style embeddings. The trained VAE and style embeddings will also be utilized in the second training stage. 

To make the style transfer focus on pivot words (e.g., emotional words in sentiment TST) while maintaining other words unchanged (so that the fluency and semantics can be largely preserved), we fine-tune the VAE with pivot word masking in training stage II. The masking is based on the probability distribution of pivot words for specific styles, which is learned from a style classification task. 


In the inference stage, VT-STOWER uses the learned external target style embeddings to adjust the sampled latent vector of the original input to the target style. The adjusted sentence vector will then be input to the decoder to generate the target style text.

\subsection{Training Stage I: VAE \& Style Embeddings}
\begin{figure*}
     \centering
     \begin{subfigure}[b]{0.45\textwidth}
         \centering
         \includegraphics[width=\textwidth]{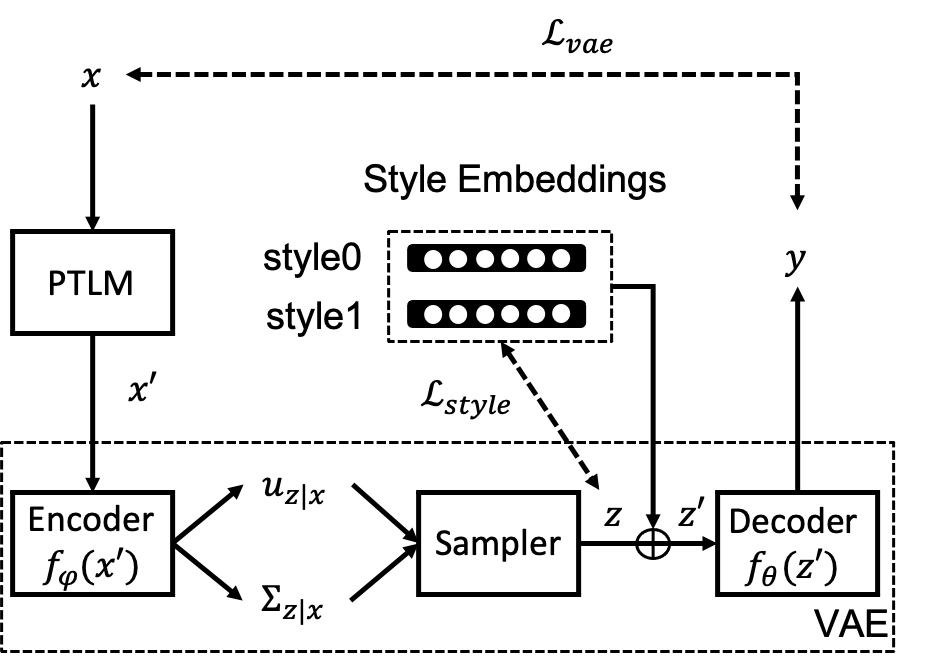}
         \caption{Training stage I: VAE \& Style Embeddings}
         \label{fig:stage1}
     \end{subfigure}
     \hfill
     \begin{subfigure}[b]{0.45\textwidth}
         \centering
         \includegraphics[width=\textwidth]{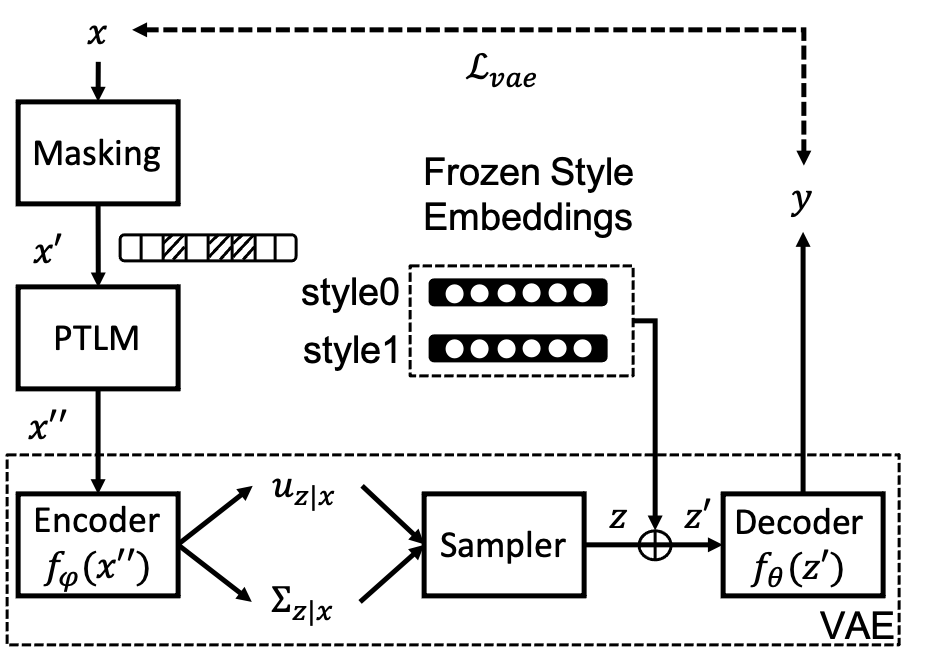}
         \caption{Training stage II: Pivot Words Enhancement}
         \label{fig:stage2}
     \end{subfigure}
     \caption{Workflow of two training stages. a) Training stage I: VAE \& style embeddings training. The VAE structure learns to reconstruct the inputs sentence $x$, and the style embeddings learn the vector representation of each style. PTLM represents Pre-Trained Language Model. b) Training stage II: pivot words masking training. The VAE is further fine-tuned with similar reconstruction task with additional pivot words masking. The frozen style embeddings are added to the latent vector to reconstruct the original sentence. We frozen the style embeddings in this step since the style-related pivot words have higher possibility to be masked and latent vector loses the style information which is the key for style embeddings training.}
     \label{fig:model}
\end{figure*}
Figure \ref{fig:stage1} presents the details of training stage I. Given a sentence $x$ whose style type is known, we firstly extract the contextualized vectors through a  pre-trained language model as the input to the VAE model, since a pre-trained language model (such as \texttt{RoBerta} \citep{liu2019roberta} and \texttt{XLM-R} \citep{conneau2020unsupervised}) can improve the performance of the downstream models,  especially when the training data size is small \citep{peters-etal-2018-deep}. After that, similar to typical VAE structure from \citet{bowman-etal-2016-generating,john-etal-2019-disentangled}, a multi-layer transformer is used as the encoder to encode $x$ to a mean vector $u \in \mathbb{R}^{d}$ and a variance vector $\Sigma \in \mathbb{R}^{d}$ to construct a latent distribution $\mathcal{N}(\mu, \Sigma)$. $d$ represents the dimension of the latent space. $z$ is the vector sampled from the latent distribution and will be input to the decoder (which is also a multi-layer transformer) to reconstruct the original text. The latent distribution is assumed to be a normal distribution $\mathcal{N}(0, \mathrm{I})$. The standard loss function of the VAE model is defined as:
\begin{equation}
    \mathcal{L}_{vae} = -\mathbb{E}_{q(z|x)}[\log p(x|z)] + \beta\cdot \mathbbm{KL}(q(z|x)\parallel p(z))
    \label{eq:vae_orig_zk}
\end{equation}
where the first term represents the likelihood of the reconstruction of the original text $x$ while the second term is the Kullback–Leibler (KL) divergence between the latent distribution and standard normal distribution. $p(z)$ represents the prior which is the standard normal distribution $\mathcal{N}(0, \mathrm{I})$, and $q(z|x)$ is the posterior distribution in the form of $\mathcal{N}(\mu, \Sigma)$. $\beta$ is the hyperparameter balancing the learning capacity between self-reconstruction and style features \citep{higgins2016beta}.

Style embeddings are also learned in this training stage. Instead of disentangling style attributes from latent features \citep{shenstyle,hu2017toward,fu2018style,john-etal-2019-disentangled}, we utilize external style embeddings to learn the original and target style representations. The advantage of external style embeddings is that they can avoid separating latent feature which leads to the lower capacity of vector representation \citep{dai-etal-2019-style}, and can differentiate the space of different styles. The set of style embeddings is defined as $S=\{s_1, s_2, \cdots, s_k\},  s_{i} \in \mathbb{R} ^{k \times d} $, where $k$ is the number of styles ($k$ is commonly to be 2 in TST tasks). Style embeddings are generated by a linear forward network whose output dimension is $d$. This style embedding network is randomly initialized and will be updated by minimizing the similarity between the style embeddings and latent feature of the input instances.

To minimize such similarity, we calculate the cosine similarity between style embeddings $s_i$ ($1\leq i\leq k$) and sampled latent feature $z$ as the style loss. The assumption is that the style embedding should be highly related to the latent feature encoded from the sentence which belongs to the same style, e.g., the distance between positive style embedding and latent vector encoded from positive sentence should be close to 1, while the distance should be 0 between positive style embedding and latent vector from the negative input sentence. The style loss is defined as follows:
\begin{equation}
    \mathcal{L}_{style} = -\sum_{i=1}^k d_i\log(\sigma(\cos(s_i, \text{sg}(z))))
    \label{eq:style_loss_zk}
\end{equation}
For brevity, we only present the loss for a single style sentence, where $d_i$ represents the ground truth distance. Specifically, if $i_{th}$ style is the style of the input sentence, $d_i=1$, otherwise, $d_i=0$. $\sigma(\cdot)$ here is the sigmoid activation function which controls the range of cosine similarity between 0 and 1. $\text{sg}(\cdot)$ is the `stop gradient' function, e.g., the feature $\text{sg}(z)$ is extracted through the latent distribution and used as an independent constant vector for computing the $\mathcal{L}_{style}$. 
The VAE loss is slightly modified from Equation \ref{eq:vae_orig_zk} by adding style embedding to hint decoder the style of sentences to be generated.
\begin{multline}
        \mathcal{L}_{vae} = -\mathbb{E}_{q(z|x)}[\log p(x|z+\text{sg}(s_x))] \\
        + \beta\cdot \mathbbm{KL}(q(z|x)\parallel p(z))
\end{multline}
where $s_x$ is the style embedding of sentence $x$. Similarly, the $s_x$ is also used as a constant vector.
Therefore, the total loss function is then defined as:
\begin{equation}
    \mathcal{L}_{total} = \lambda_{vae}\mathcal{L}_{vae}  +\lambda_{style}\mathcal{L}_{style} 
    \label{eq:total_loss}
\end{equation}
where $\lambda_{vae}$ and $\lambda_{style}$ are penalty weights, which are hyperparameters to balance between VAE loss and style embeddings loss.

\subsection{Training Stage II: Pivot Words Masking}

In Stage I, we co-train VAE \& style embeddings where we show how to leverage learned style embeddings to further improve the VAE model. In stage II, We further enhance the model by masking pivot words to prompt decoder to focus on pivot words, because certain style-related words play crucial roles in TST \citep{fu2019rethinking}. For instance, the pivot word of the sentence `I am disappointed with the restaurant' in sentiment transfer is `disappointed' because this word contributes the most to the negative sentiment. However, other words such as "I, was" are anchor words, which are unrelated to the sentiment but affect the semantics thus should be unchanged during the style transfer. Therefore, this stage of training is important to enhance the model ability in transferring pivot words while keeping anchor words. This stage cannot be merged into training stage I because 1) in this stage style embeddings have no visibility to the style-related pivot words so that the style information is hard to be learned; 2) the style embeddings learned in training stage I have auxiliary function in helping reconstructing masked pivot words during fine-tuning the VAE.

However, randomly masking words in input sentence and only relying on style embeddings to emphasize the pivot words does not achieve ideal results. A more efficient way is to learn which words are more possible to be pivot words for a specific style, and mask them based on the probability. Similar to \citet{sudhakar2019transforming}, we utilizes the importance score distribution to 
indicate the possibility of words being pivot (a pivot word has a higher score). Such importance score distribution is achieved from the attention weights of a style classifier. Specifically, we train a style classifier based on a pre-trained language model, appending with a softmax layer over the attention stack of the first token. The first token is usually a special symbol that represents the beginning of the sentence (e.g., `$<$s$>$'), and also collects other tokens' attention weights that correspond to their significance in identifying the style of the input sentence. The importance score of a token $w$ in the input sentence $x$ is defined as follows:
\begin{equation}
    \alpha(w) = \frac{1}{L}\sum_{i=1}^L\text{softmax}_{w\in x}(\frac{Q_{<s>, i}K_{w, i}^T}{\gamma})
    \label{eq:scores}
\end{equation}
where $L$ is the number of attention heads. $Q, K$ are quires and keys in the final layer of the language model \citep{vaswani2017attention}. Their subscript $<w, i>$ represents the vector of token $w$ in $i_{th}$ head. $\gamma$ is a hyperparameter ranging in (0,1) to adjust the sharpness of the score distribution (smaller means sharper).

After we get the pivot words probability, we mask words in the input sentences based on this importance score distribution. Specifically, every token $x_i$ is assigned a random number $p_i$, conforming to the uniform distribution $p_i\sim\text{uniform[0,1]}$. Tokens are masked into a special symbol `$<$mask$>$' if their assigned number is smaller than the score ($p_i<\alpha(x_i)$) so that words that possess higher important scores have higher probability to be masked. Following the previous example, the input sentence would be masked as `I was $<$mask$>$ with the restaurant'. In this way, masked sentence preserves the content but with style attributes removed. Then the VAE model is fine-tuned to reconstruct masked sentence to the original sentence by adding the corresponding style embedding to the latent feature. The loss function is defined as follows:
\begin{multline}
    \mathcal{L}_{vae} = -\mathbb{E}_{q(z|x)}[\log p(x|z+\text{sg}(s'_x))] \\
    + \beta\cdot \mathbbm{KL}(q(z|x)\parallel p(z))
    \label{eq:vae_mask_zk}    
\end{multline}
where $s'_x$ is the style embedding which has the same style as input $x$. Note that we do not update style embeddings in this stage because the style embeddings are used for assisting fine-tuning the decoder with style information, and their general style representation should not be impacted by the pivot words reconstruction loss. Moreover, to prevent the latent space of VAE from shifting or distorting to unreasonable distribution that only describes masked sentences, we conduct pivot words masking in randomly 50\% of sentences.

Although \citet{madaan-etal-2020-politeness} has a similar method tagging the source style phrases and generating the target style sentences by using n-gram tf--idfs, the core differences of our stage II method are: 1) each word has a probability of being masked calculated by the attention scores of the stacked classifier on a language model, which leads to a smooth word masking probability distribution; 2) VAE decoder reconstructs the masked sentences using both information of latent space and external style embeddings.

\subsection{Inference stage}
In the inference stage, the latent representation z generated from the input sentence $x$ through VAE will be adjusted before sending to the decoder. In detail, the latent vector z will be added the target style embedding and subtracted the style embedding of original style as $x$. Intuitively, we expect the injection and removal of style information is completed by the addition and subtraction operations of style embeddings. The updated latent representation is expressed as follows:
\begin{equation}
    z' = z + w\cdot(s_t - s_o)
    \label{eq:inference}
\end{equation}
where $s_t$ and $s_o$ are target and original style embeddings trained in the stage I respectively. $w$ represents the style weight that adjusts the style strength applied to the sentence generation. A higher weight means stronger style attributes will be injected for generation.

\section{Experiments}
\subsection{TST Evaluation Tasks and the Settings}
We evaluate VT-STOWER with three different TST tasks: sentiment transfer, formality transfer, and code-switching transfer. 

For sentiment transfer, we adopt the Yelp dataset \citep{li-etal-2018-delete}, in which each sample is a business review of a restaurant and is labeled as positive or negative.
For formality transfer, We adopt one of the largest corpus for formality transfer task, namely \textit{Family and Relationships} domain data in GYAFC (Grammarly’s  Yahoo Answers  Formality  Corpus) \citep{rao-tetreault-2018-dear}. For code-switching transfer, we evaluate VT-STOWER on a Hindi$\rightarrow$Hinglish transfer task, which is extracted from the English-Hinglish translation dataset at LinCE (Linguistic Code-switching Evaluation) \citep{aguilar-etal-2020-lince}. We first translate English sentences into Hindi by Amazon Translation Service and then transliterate Latin scripts of Hindi words into Devanagari form by using \textit{indic-trans} tool \citep{10.1145/2824864.2824872} to keep the consistency of the script of language \footnote{The output of translator is Devanagari form while the original script of Hindi in LinCE is Latin.}. Note this dataset is very low-resource, which only contains 7K sentences for training. Similar to GYAFC set, we shuffle the training data and treat it as unpaired data. Note that two of the training sets are transformed from originally paired dataset instead of directly using richer unpaired datasets, it is because we want to make a fair comparison with other referenced approaches. The training and test set size for each task is presented in Table \ref{tab:dataset-size}.
\begin{table}[ht]
\resizebox{1\linewidth}{!}{
\begin{tabular}{lcccc}
\multicolumn{1}{l|}{Tasks}                 & training set   & evaluation set    & test set \\ \hline
\multicolumn{1}{l|}{Sentiment Transfer (positive/negative)}         & 266K/177K  & 2K/2K  & 500/500         \\
\multicolumn{1}{l|}{Formality Transfer (formal/informal)}              & 52K/52K  & 2.2K/2.7K  & 1K/1.3K          \\
\multicolumn{1}{l|}{Code-Switching Transfer (Hinglish/English)}     & 7K/7K  & 300/300  & 300/300  \\
\hline
\end{tabular}
}
\caption{Training, evaluation, and test set size of three evaluation tasks.}
\label{tab:dataset-size}
\end{table}

Considering that GYAFC and Yelp dataset are written in English and the code-switching dataset is in mixed languages of Hindi and English, we use \texttt{RoBERTa} as the pre-trained language model for sentiment and formality transfer tasks, and \texttt{XLM-R} \citep{conneau2020unsupervised} for code-switching transfer. Also, we fine-tune the style classifier to obtain important score distribution by leveraging \texttt{RoBERTa} and \texttt{XLM-R} for the corresponding transfer tasks. More training hyperparameters are shown in Appendix \ref{app:hyperparamerters}.

\subsection{Evaluation Metrics}
\paragraph{Style Transfer Accuracy (Acc)}
Style transfer accuracy (Acc) is defined as the ratio of the number of successfully transferred sentences and the total number of input sentences. Following previous studies \citep{dai-etal-2019-style,sudhakar2019transforming}, we leverage fastText classifier \citep{joulin-etal-2017-bag} to classify whether the original text have been successfully transferred to the target style. The classifier is trained on the same training data used for style transfer. The three classifiers achieve 97.6\%, 85.75\% and 99.7\% accuracy for sentiment, formality and code-switching style classification itself, respectively.
\paragraph{Perplexity (PPL)}
We also measure the fluency of the transferred sentences by calculating their perplexity. The lower the perplexity is, the more fluent the generated sentences are. For the  GYAFC and Yelp dataset which are in English, we use the pre-trained language model GPT2 \citep{radford2019language} to compute the perplexity, where no further fine-tuning is conducted. However, GPT2 does not apply to other languages or code-switching sentences. Following \citet{samanta2019deep}, we train a character-level LSTM \citep{kim2016character} on the code-switching training data and utilize this model to derive the perplexity of generated code-switching sentences.
\paragraph{BLEU Scores}
Content preservation is evaluated by the tokenized BLEU scores \citep{papineni2002bleu} between the transferred sentences and human-authored references, which is calculated with the \texttt{multi-bleu.perl}. Note that GYAFC dataset has four human references, so the BLEU for GYAFC is the mean BLEU scores between the generated sentences and four references. Because there is no human reference for code-switching task, we report BLEU scores between transferred sentences and original sentences for code-switching transfer instead.
\paragraph{Geometric Mean (GM)}
Following \citet{ijcai2020-526}, We also report the geometric mean of accuracy,
BLEU, $\frac{1}{\ln \text{PPL}}$ as the overall performance.

\begin{figure*}[ht]
     \centering
     \begin{subfigure}[b]{0.32\textwidth}
         \centering
         \includegraphics[width=\textwidth]{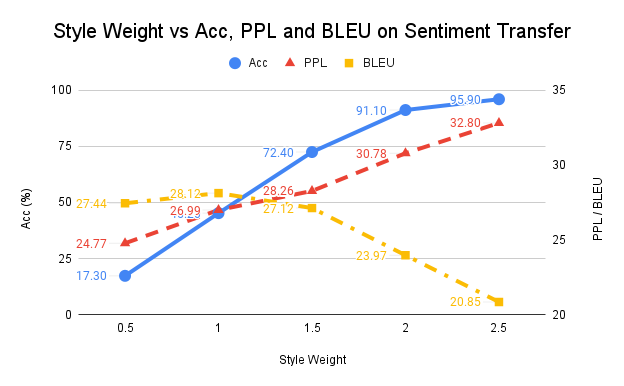}
         \caption{sentiment transfer}
         \label{fig:style_weights:sentiment}
     \end{subfigure}
     \hfill
     \begin{subfigure}[b]{0.32\textwidth}
         \centering
         \includegraphics[width=\textwidth]{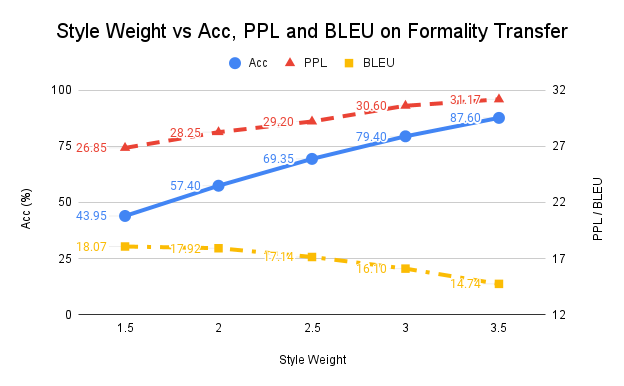}
         \caption{formality transfer}
         \label{fig:style_weights:formality}
     \end{subfigure}
    \hfill
     \begin{subfigure}[b]{0.32\textwidth}
         \centering
         \includegraphics[width=\textwidth]{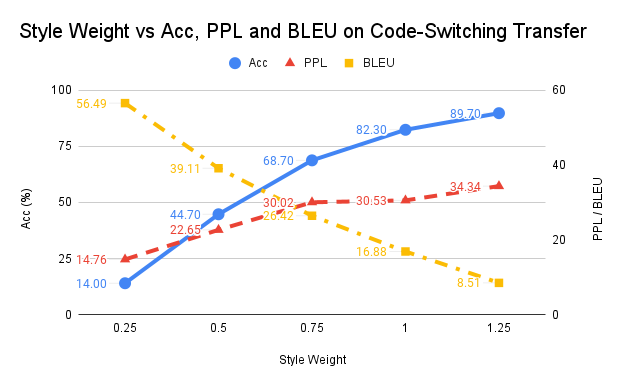}
         \caption{code-switching transfer}
         \label{fig:style_weights:cs}
     \end{subfigure}
     \caption{Illustration of style weight $w$ vs. Acc, PPL and BLEU in sentiment, formality and code-switching transfer tasks. Note there is a trade-off between Acc and PPL/BLEU. With increasing of $w$, Acc will increase while BLEU drops down and PPL increases.}
     \label{fig:style_weights}
\end{figure*}

\begin{table}[ht]
\resizebox{1\linewidth}{!}{
\begin{tabular}{lcccc}
\multicolumn{1}{l|}{Models}                 & Acc $\uparrow$   & PPL $\downarrow$  & BLEU $\uparrow$ & GM $\uparrow$ \\ \hline
\multicolumn{5}{c|}{Sentiment Transfer (Yelp)}                                      \\ \hline
\multicolumn{1}{l|}{CrossAlignment}         & 74.0  & 42.91   & 9.06   & 5.63 \\
\multicolumn{1}{l|}{Delete \& Retrieve}     & 87.5  & 40.66 & 5.99  & 5.21  \\
\multicolumn{1}{l|}{B-GST}                  & 84.3  & \bf 25.27  & 22.82  &  8.41 \\
\multicolumn{1}{l|}{Style Transformer}      & 83.9  & 43.60  & \bf 28.29 &   8.57 \\
\multicolumn{1}{l|}{Deep LatentSeq}         & 83.0  & 27.08  & 24.03 & 8.46 \\
\multicolumn{1}{l|}{StyIns}                 & 91.5  & 42.60  & 25.11  & 8.49\\
\multicolumn{1}{l|}{Tag \& Generate}         & 87.5 & 32.98   & 21.80 & 8.17 \\
\multicolumn{1}{l|}{Ours (stage I, $w=4$)}  & \bf 91.7  & 38.35   & 18.51  & 7.75 \\
\multicolumn{1}{l|}{Ours (stage II, $w=2$)}  & 91.1  & 30.78  & 23.97  &  \bf 8.61\\\hdashline
\multicolumn{1}{l|}{Human Reference}        & 74.1  & 27.40   & 100.0  &  13.08 \\ \hline
\multicolumn{5}{c}{Formality Transfer (GYAFC)}                                      \\ \hline
\multicolumn{1}{l|}{CrossAlignment}         & 65.35 & \bf 13.66  & 1.57 & 3.40    \\
\multicolumn{1}{l|}{Delete \& Retrieve}     & 53.85 & 29.70   & 11.71   & 5.71 \\
\multicolumn{1}{l|}{Style Transformer}      & 56.05 & 48.72   & \bf 24.67  & 7.09 \\
\multicolumn{1}{l|}{Ours (stage I, $w=4$)}    & 80.9 & 31.90  & 14.19  & 6.92\\
\multicolumn{1}{l|}{Ours (stage II, $w=3.1$)}  & \bf 81.0  & 30.78   & 15.84 &\bf 7.21\\\hdashline
\multicolumn{1}{l|}{Human Reference}        & 82.31 & 28.05  & 100.0  & 13.39\\ \hline
\multicolumn{5}{c}{Code-Switching Transfer (LinCE)}                                 \\ \hline
\multicolumn{1}{l|}{Style Transformer}      & \bf 99.3  & 601.45 & 3.47      & 3.78       \\
\multicolumn{1}{l|}{Randomly Replace}       & 1.02	& 213.24 & \bf 69.09	& 2.36    \\
\multicolumn{1}{l|}{Ours (stage I, $w=0.75$)} & 66.67 & 29.91  & 24.30     & 7.81       \\
\multicolumn{1}{l|}{Ours (stage II, $w=0.75$)} & 68.70 & 30.02  & 26.42     & \bf 8.11       \\ \hline
\end{tabular}
}
\caption{Overall results of our models (VT-STOWER) and previous methods on three style transfer tasks. The best scores are bolded in the corresponding metric. $\uparrow$ means the higher is better, vice versa.}
\label{tab:main-result}
\end{table}

\begin{figure}[ht]
    \centering
    \includegraphics[width=0.48\textwidth]{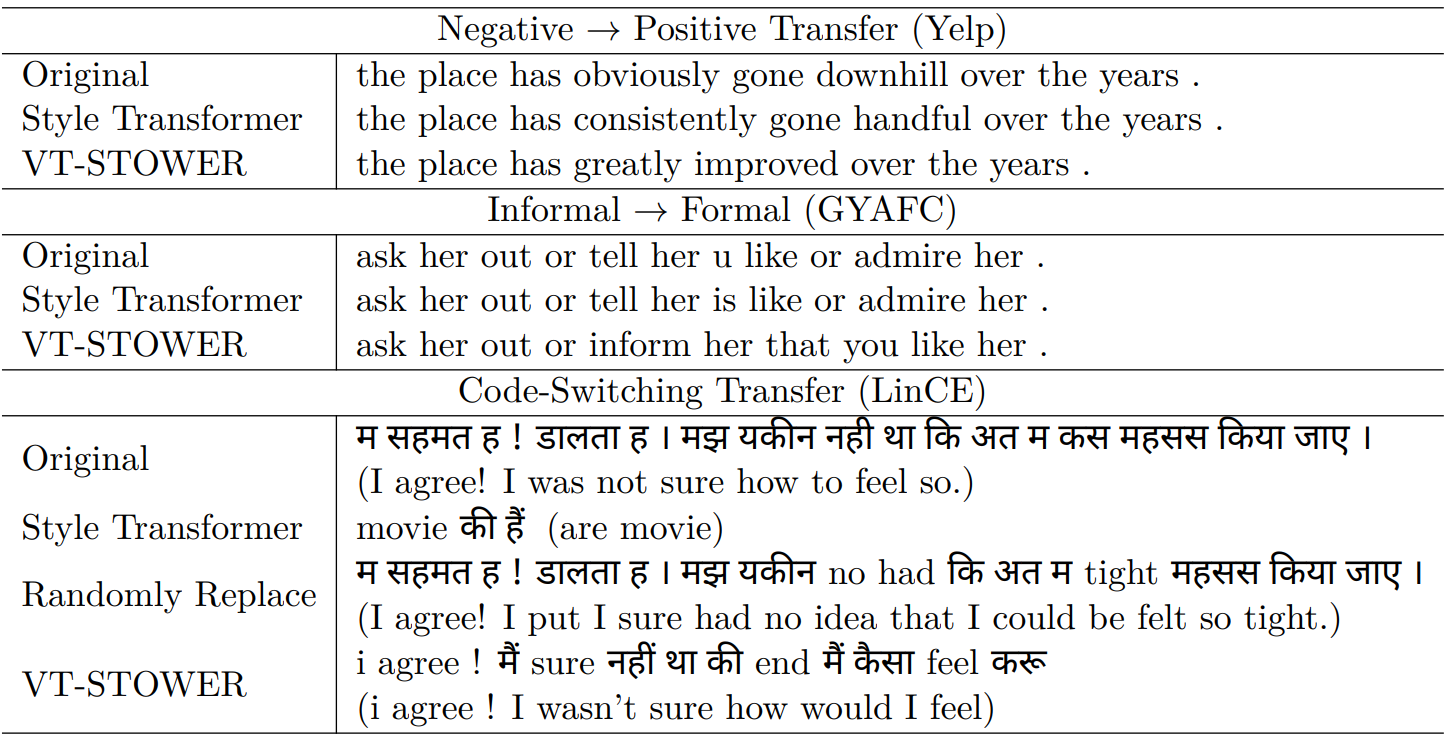}
    \caption{Comparison of style transfer outputs of our models and style transformer in three transfer tasks. Our models are stage II models in Table \ref{tab:main-result}. Translations for code-switching sentences are shown in parenthesis.}
    \label{fig:examples}
\end{figure}

\begin{figure*}[ht]
     \centering
     \begin{subfigure}[b]{0.27\textwidth}
         \centering
         \includegraphics[width=\textwidth]{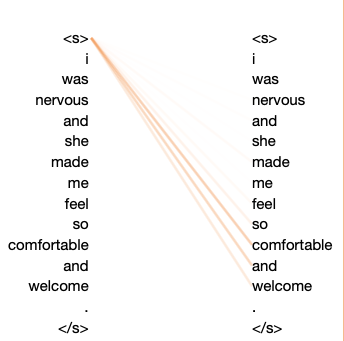}
         \caption{sentiment transfer}
     \end{subfigure}
     \hfill
     \begin{subfigure}[b]{0.27\textwidth}
         \centering
         \includegraphics[width=\textwidth]{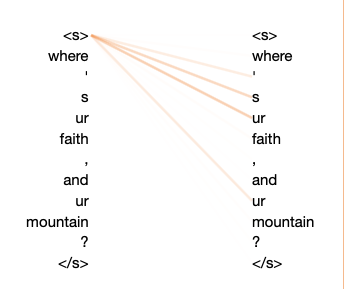}
         \caption{formality transfer}
     \end{subfigure}
    \hfill
     \begin{subfigure}[b]{0.27\textwidth}
         \centering
         \includegraphics[width=\textwidth]{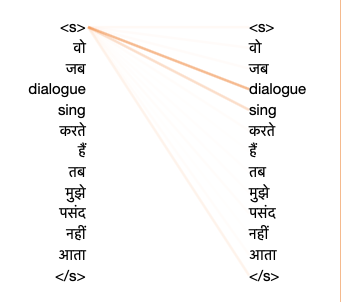}
         \caption{code-switching transfer}
     \end{subfigure}
     \caption{Illustration of the mean attention weights of token `$<$s$>$' from all heads at the final layer in three TST tasks. Higher importance scores are assigned to pivot words, which are depicted as deeper lines in the figures.}
     \label{fig:importance}
\end{figure*}

\subsection{Main Results}
\label{sec:main_result}
The performance of VT-STOWER and previous works are shown in Table \ref{tab:main-result}. First of all, we can clearly see the performance improvement brought by stage II training compared with single training stage I. \textbf{In all three transfer tasks, models trained in stage II lead to lower PPL (or similar PPL in code-switching transfer) and higher BLEU scores when we find a $w$ to control them in a similar Acc, which achieves better overall performance.} For instance, compared with the model trained in stage I with $w=4$ in the sentiment transfer, the model fine-tuned in stage II achieves similar accuracy with $w=2$ (91.7\% vs. 91.1\%). At the same time, the stage II model decreases the PPL from 38.35 to 30.78 and increases the BLEU from 18.51 to 23.97, which demonstrates that the pivot words masking training is capable of improving the smoothness of the sentences and the preserving the content. Note that we cannot use the same weight $w$ for a direct comparison since the models in two training stages have different sensitivity to $w$. Therefore, we use $w$ that produces similar accuracy between stage I and stage II for a fair comparison for the PPL and BLUE.

When comparing our method with several state-of-the-art references: CrossAlignment \citep{shenstyle}, Delete \& Retrieve \citep{li-etal-2018-delete}, B-GST \citep{sudhakar2019transforming}, Style Transformer \citep{dai-etal-2019-style}, Deep LatentSeq \citep{He2020A}, Tag \& Generate \citep{madaan-etal-2020-politeness} and StyIns \citep{ijcai2020-526}, the performance of their methods is directly evaluated on their provided outputs by using our metric evaluators. We will further discuss how $w$ affect the performance in next section. We can clearly see the overall performance (GM) of our proposed model is better than all baselines. For evaluating the success of style transfer, accuracy is the most critical metrics, for which VT-STOWER also demonstrates large improvement in sentiment and formality transfer.

In the sentiment style transfer, our model with $w=2$ (after stage II training) has competitive accuracy (91.1\%), and BLEU (23.97) compared with the state-of-the-art methods StyIns \citep{ijcai2020-526} (accuracy=91.5\%, BLEU=25.41) and style transformer \citep{dai-etal-2019-style} (accuracy=83.9\%, BLEU=28.29) but achieve much lower perplexity (30.78) compared to 42.60 in StyIns and 43.60 in style transformer, which demonstrates that the sentences generated from our model is closer to the natural language. VT-STOWER also outperforms other previous methods by a large margin in all three metrics. 
In the formality transfer, the most competitive model is the style transformer. Although it achieves higher BLEU scores (24.67), our models beats it on higher style transfer accuracy (81.0\% vs. 56.05\%) and significantly lower PPL (30.78 vs. 48.72) with limited loss of BLEU scores. 

For the code-switching transfer, since there is no previous TST experimenting on this task, we train the strongest baseline (style transformer) for this task. Interestingly, style transformer obtains a very high accuracy (99.8\%) with the costs of very high PPL (601.46) and very low BLEU score (3.47). The possible reason is that the style transformer is only able to capture partial special style features from the small dataset (7K) and only transfer sentences based on these features without fully capturing the nature of languages, resulting in high accuracy but low fluency and BLEU. However, VT-STOWER can balance among the accuracy, fluency, and BLEU to achieve reasonable results even in the case of the low-resource dataset, which demonstrates its generalization power. Additionally, we also design another baseline, i.e., we randomly replace 15\% Hindi words with English words \citep{zheng2021consistency} based on the MUSE dictionary \citep{conneau2017word}, because intuitively, people may regard code-switching text generation as simply translating several words. However, this method only achieve 1.02\% accuracy, because simple translation and replacement cannot accord with the habit of bilingual expression in code-switching sentences, namely, code-switching has its own style according to the speakers (e.g., usually noun is more likely to be replaced with foreign language than preposition). Intuitively, when we compare the VT-STOWER with original and other approaches as shown in Figure \ref{fig:examples}, the output bilingual sentence from VT-STOWER reads more fluent and can be easier understood.

\subsection{Effect of Style Weights}
\label{sec:effect_style_weights}
As shown in Equation \ref{eq:inference}, the strength of the target style in $z'$ is adjusted by the style weight $w$. In Figure \ref{fig:style_weights}, we present metrics trend with five different $w$ for the models trained in stage II \footnote{$w$ ranges from 0.5 to 2.5 with an interval of 0.5 for sentiment and formality transfer, and from 0.25 to 1.25 with an interval of 0.25 for code-switching transfer.}, and demonstrate how the style weight $w$ affects the outputs. Taking sentiment transfer task as an example, when $w$ is increased from 0.5 to 2.5, the transfer accuracy climbs from 17.3\% up to 95.9\%, but the BLEU score drops from 27.44 down to 20.85, and PPL increases from 24.77 to 32.8. The reason is when increasing $w$, more style information is injected into the latent vector that the decoder pays more attention to the target style feature rather than the naturalness and content of generated sentences. Therefore, $w$ is a trade-off hyperparameter between the transfer accuracy and PPL/BLEU. Examples of generated sentences transferred from positive to negative sentiment with $w=1.5,2,2.5$ are illustrated in Table \ref{tab:style-weights}. When $w=1.5$, the model still can find a positive word, `enjoying,' which makes the sentence ironical. In the case of $w=2$, the `enjoying' is rephrased to `avoid', turning the sentence into a full negative attitude. If we further increment $w=2.5$, more negative words will be added regardless of the smoothness of the sentence. Similar discussions also hold for the formality and code-switching transfer, where their results versus various style weights are illustrated in Figure \ref{fig:style_weights:formality} and  \ref{fig:style_weights:cs}.

\begin{table}[]
\resizebox{1\linewidth}{!}{
\begin{tabular}{ll}
\hline
\multicolumn{2}{c}{Positive $\rightarrow$ Negative with various $w$}                      \\ \hline
\multicolumn{1}{l|}{Original} & i will be going back and enjoying this great place ! \\ \hdashline
\multicolumn{1}{l|}{$w=1.5$}  & i will be going back and enjoying this \textbf{terrible} place ! \\
\multicolumn{1}{l|}{$w=2$} & i will be going back and \textbf{avoid} this \textbf{terrible} place ! \\
\multicolumn{1}{l|}{$w=2.5$}  & i will be going back and \textbf{worst rude avoid} this \textbf{terrible} place ! \\ \hline

\end{tabular}}
\caption{Examples of sentences transferred from positive to negative sentiment with various settings of $w$. The higher $w$ is the more negative words are injected in the sentences.}
\label{tab:style-weights}
\end{table}

\subsection{Importance Score Distribution}
Recall that for the training stage II, the importance scores are derived from the attention scores of the BOS token `$<$s$>$', which are the mean scores of all heads from the last layer of a pre-trained encoder. Figure \ref{fig:importance} presents the examples of importance score, showing how the score value represents the importance of words in terms of style representation. The importance scores are higher on `comfortable and welcome' in the sentiment transfer, these words represents strong positive emotions. Similarly, the scores are higher on the informal written words `ur' in the formality transfer, and English words mixed in a Hinglish sentence in the code-switching transfer.

\section{Conclusion}
We proposed the VT-STOWER, a model joinly trained with VAE and style embeddings for content distribution and style information. The method successfully transfers several different text styles, including the code-switching TST task for the first time. Taking advantage of the flexibility of style embeddings, our proposed model has the ability to adjust the style strength during the transfer by simply adjusting the style weights. To further enhance the transfer accuracy, we propose additional pivot words masking training scheme, which shows impressive improvement.



\bibliography{anthology,acl2020}
\bibliographystyle{acl_natbib}
\clearpage
\appendix
\section{Hyperparamerters}

\label{app:hyperparamerters}
The encoder and decoder for sentiment and formality transfer tasks both are two-layer transformer \citep{vaswani2017attention}, with FFN dimension size of 1024 and 4 attention heads. Due to the limited code-switching data size, we run smaller encoder and decoder with 256 FFN dimensions and 2 attention heads for code-switching transfer task. The neural networks that formulate the mean and variance of the latent space are also one-layer transformer blocks. The dimension of the latent features and style embeddings is 768. Both penalty weights $\lambda_{vae}$ and $\lambda_{style}$ equal to 1.  We set $\gamma$ as 0.01, 0.03, and 0.005 for sentiment, formality, code-switching transfer during importance score calculation, respectively. The $\beta$ is set as 1. The optimizer is Adam \citep{kingma2014adam} with learning rate 0.0005. The batch size is 8092 tokens.





\end{document}